\documentclass[letterpaper]{article} 
\usepackage{aaai25}  
\usepackage{times}  
\usepackage{helvet}  
\usepackage{courier}  
\usepackage[hyphens]{url}  
\usepackage{graphicx} 
\usepackage{natbib}  
\usepackage{caption} 
\usepackage{multirow}
\usepackage{amsmath}
\usepackage{xcolor}
\usepackage{algorithm}
\usepackage{algorithmic}

\usepackage{newfloat}
\usepackage{listings}
\DeclareCaptionStyle{ruled}{labelfont=normalfont,labelsep=colon,strut=off} 
\floatstyle{ruled}
\newfloat{listing}{tb}{lst}{}
\floatname{listing}{Listing}
\title{Speech Representation Learning Revisited: The Necessity of Separate Learnable Parameters and Robust Data Augmentation}
\author{
   \Large { 
    Hemant Yadav \textsuperscript{\rm 1}, 
    Sunayana Sitaram \textsuperscript{\rm 2}, 
    Rajiv Ratn Shah \textsuperscript{\rm 1} 
    } 
}
\affiliations{
    \Large {
    \textsuperscript{\rm 1} IIIT Delhi, India, 
    \textsuperscript{\rm 2} Microsoft Research Banglore, India 
    } \\
   \large {
   \{hemantya,rajivratn\}@iiitd.ac.in, sunayana.sitaram@microsoft.com 
   }
}

\usepackage{bibentry}

\begin{document}

\maketitle

\begin{abstract}
Speech modeling methods learn one embedding for a fixed segment of speech, typically in between 10-25 ms. 
The information present in speech can be divided into two categories: "what is being said" (content) and "how it is expressed" (other) and these two are orthogonal in nature causing the optimization algorithm to find a sub-optimal solution if forced to optimize together. This leads to sub-optimal performance in one or all downstream tasks as shown by previous studies. 
Current self-supervised learning (SSL) methods
such as HuBERT are very good at modeling the content information present in speech. Data augmentation improves the performance on tasks which require effective modeling of other information but this leads to a divided capacity of the model.
In this work, we conduct a preliminary study to understand the importance of modeling other information using separate learnable parameters. We propose a modified version of HuBERT, termed Other HuBERT (O-HuBERT), to test our hypothesis.
Our findings are twofold: first, the O-HuBERT method is able to utilize all layers to build complex features to encode other information; second, a robust data augmentation strategy is essential for learning the information required by tasks that depend on other information and to achieve state-of-the-art (SOTA) performance on the SUPERB benchmark with a similarly sized model (100 million parameters) and pre-training data (960 hours).

\end{abstract}

\section{Introduction}
\label{section:intro}
Self-supervised learning (SSL) techniques have gained significant attention in recent years for their ability to learn high-level representations from speech data \cite{baevski2020wav2vec2.0, hsu2021hubert, chen2022wavlm, baevski2022data2vec, chung2021w2vbert}. Depending on the pretext task used during pre-training, SSL models can be categorized into three distinct groups: (i) models employing masked predictive coding (MPC) as seen in \cite{liu2020mockingjay, hsu2021hubert, chen2022wavlm}, (ii) models using auto-regressive predictive coding (APC) exemplified by works like \cite{brown2020languagegpt3, chung2019unsupervised, chung2020generative, ling2020decoar2}, and (iii) contrastive predictive coding (CPC), demonstrated in \cite{oord2018representationcpc, vqw2v, schneider2019wav2vec, baevski2020wav2vec2.0}. 
These learned representations can then be used as input features for various downstream tasks such as understanding the content (i.e., what is being said), and other (how it is expressed, closely tied to aspects like speaker characteristics and para-linguistics). 

SUPERB \cite{yang2021superb} is one such benchmark to study the generalization capabilities of different speech representation learning methods by evaluating the learned representations on a variety of downstream tasks such as automatic speech recognition (ASR), phoneme recognition (PR), speaker identification (SID), emotion recognition (ER), and voice conversion (VC).
State-of-the-art (SOTA) methods \cite{chen2022wavlm, hsu2021hubert, wang2022lighthubert, baevski2022data2vec} on SUPERB are all modelled using a common token and masked prediction loss to encode different information types present in speech essential for downstream tasks. This modeling strategy has one significant flaw: the two types of information present in speech, content and other, are orthogonal in nature, which makes joint optimization difficult. Perfect features for an ASR system should only encode content information i.e., what is being said and forget about everything other  i.e., how it is expressed.
As a result, the model is forced to divide its encoder capacity to encode both types of information, with later layers focusing on encoding content information and earlier layers on other information. The distribution of layers depends on the specific modeling strategy employed. For example, the authors of \cite{yadav2023analysing} demonstrated that the masked prediction loss (MPL) used in HuBERT and content information are positively correlated, leading layers closer to where MPL loss is applied to encode information relevant to ASR and PR tasks. The authors of UniSpeech-SAT \cite{chen2022unispeechsat} showed that incorporating data augmentation during pre-training is essential to increases the number of layers dedicated to encoding other (speaker) information. This lead to the encoder dividing its capacity between content and other and therefore reduction in performance on ASR and PR task. The authors of WavLM \cite{chen2022wavlm} added gated relative position bias \cite{chi2021xlmwavlmgated} to mitigate the performance drop on ASR tasks.However, because they also used a common token to encode both content and other information, the model's capacity was divided between these two types of information.

The core idea of this work is to explore the importance of encoding other information using separate learnable parameters, which differentiates this study from previous works such as UniSpeech-SAT \cite{chen2022unispeechsat} and WavLM \cite{chen2022wavlm}.
Our contributions in this paper can be summarized as follows:

\begin{enumerate}

\item We present a modified version of the original HuBERT model \cite{hsu2021hubert}, specifically designed to model/encode/learn other (O) information using separate learnable parameters. This modified model is referred to as Other HuBERT (O-HuBERT). 

\item Recognizing that loss functions guide the nature of learned information \cite{yadav2023analysing, mohamed2022self}, we introduce a new loss function to maximize the other information during pre-training. This approach differs from \cite{chen2022unispeechsat} in that our loss function is optimized to train a separate token specifically for modeling other information.

\item Unlike UniSpeech-SAT \cite{chen2022unispeechsat} and WavLM \cite{chen2022wavlm}, we apply two-stage data augmentation to O-HuBERT, which involves utterance mixing followed by reverberation. Thanks to this robust data augmentation strategy, O-HuBERT achieves a new SOTA on the SUPERB benchmark when compared to similarly sized models (~100 million parameters) and pre-training data (~1000 hours).
    
\end{enumerate}

The paper is organized as follows: First, we explain the methodology employed in our study, the O-HuBERT model. Next, we outline the experimental setup and data utilized for evaluation. Following that, we present the results of our experiments. We then delve into related works, providing a comprehensive overview of existing approaches in the field. Finally, we conclude with our findings and outline potential avenues for future research.

\section{Methodology}
\label{section:method}
\begin{figure*}[ht]
\centering
\includegraphics[trim=10.5cm 5cm 10.5cm 1.5cm, clip,width=2\columnwidth]{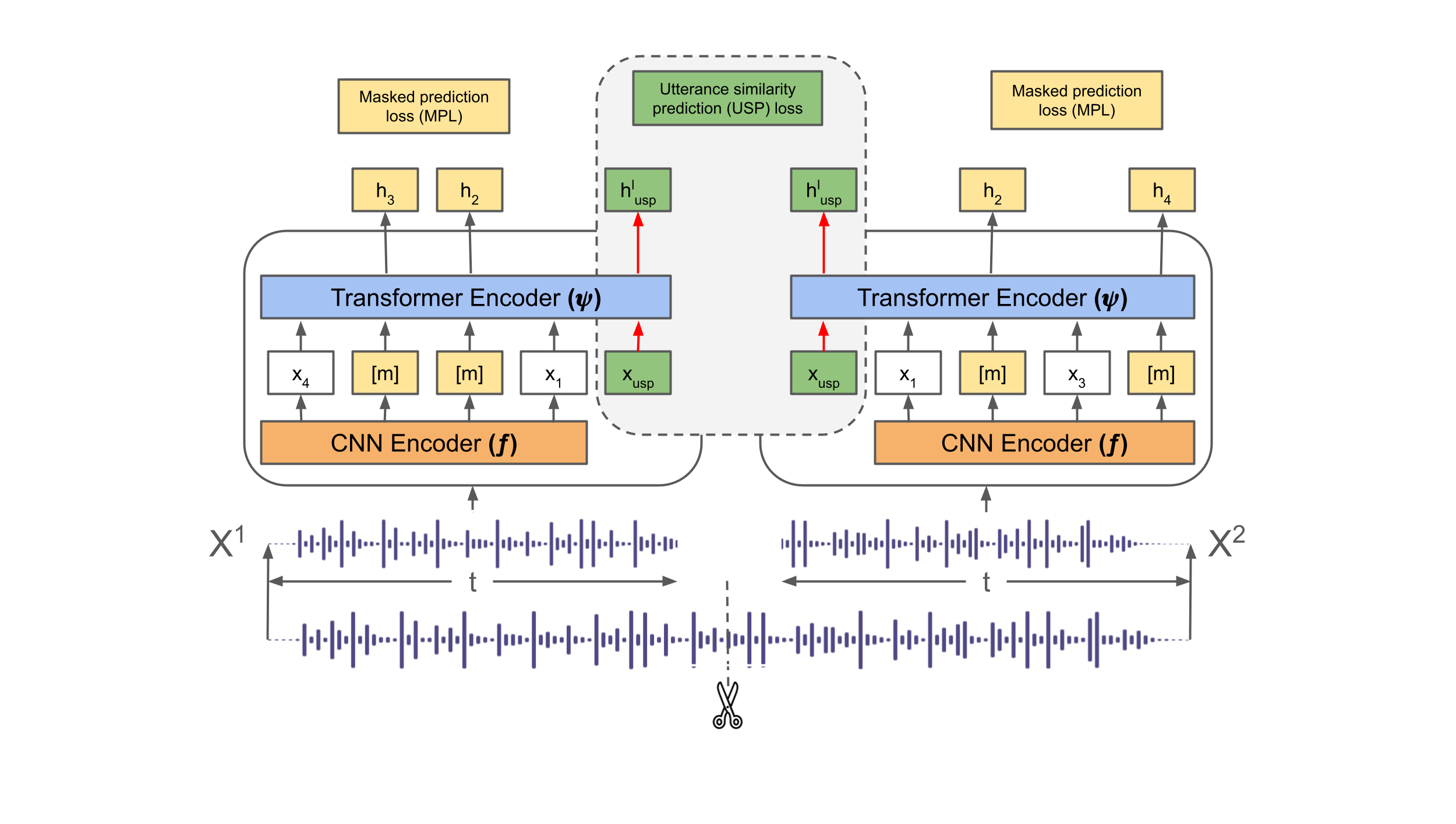}
\caption{The figure illustrates the processing of two speech utterances in a batch. In our proposed O-HuBERT model, content information is encoded using masked token indices, while other information is modeled using the USP token. Notably, calculating the other information loss involves pairs of samples from the batch, whereas the content loss does not require such pairs.}
\label{figure:proposed method}
\end{figure*}

In this section we explain the O-HuBERT model for learning speech representations, to model the other (O) and content (C) information jointly using separate learnable parameters during pre-training. We begin by explaining the process of modeling content (C) information followed by modeling other (O) information and their loss functions respectively.

\subsection{Modeling the Content Information}

To model the content (C) information we use HuBERT model as shown in Figure \ref{figure:proposed method}. Yadav et al. \cite{yadav2023analysing} showed that the masked prediction loss maximizes the content information learned during pre-training. 
HuBERT, as introduced by \cite{hsu2021hubert}, learns high level features to model the content information present in the raw speech waveform.
It is an iterative pre-training SSL model comprising of two encoders i.e, a convolution neural network (CNN) feature encoder ($f$) followed by a  transformer encoder ($\psi$). The CNN encoder serves the dual purpose of down-sampling the input data and as an embedding layer to the transformer encoder. The resulting output is passed, denoted as $U$, to the transformer encoder and is used to calculate the masked prediction loss. 
Briefly, during the pre-training stage, raw speech waveform is passed to the CNN encoder followed by masking, using the masking token $[M]$, approximately $50\%$ of the output embeddings before passing to the transformer encoder.

The objective is to minimize the MPL to output a discrete target sequence, guided by pseudo labels.
The complete details can be found in the original HuBERT paper \cite{hsu2021hubert}.

\subsection{Modeling the Other Information}

To model the other (O) information, similarly to the Next Sentence Prediction (NSP) token in the original BERT \cite{devlin2018bert} paper, we propose to append an utterance similarity prediction token ($X_{USP}$), at the beginning, to the input embeddings of transformer encoder ($\psi$) as shown in Figure \ref{figure:proposed method}. This increases the total sequence length by $1$. This completes one pass of the O-HuBERT model as depicted in Figure \ref{figure:proposed method}. 

The objective here is to classify whether two USP token embeddings are originating from the same audio utterance or not. See Section \ref{sec:method-loss} for the formulation of the proposed loss.

\subsection{Loss Formulation}
\label{sec:method-loss}

During the pre-training stage, an audio utterance ($\boldsymbol{X_i}$) in a batch of size $n$, consisting of $2t$ time steps, is split into two equal segments. Where the first segment is called key and the latter is denoted as query resulting in two utterance of $t$ time steps each as shown in Figure \ref{figure:proposed method}. This splitting the input strategy, is inspired by the work of \cite{audiodivide}, improves the learning of effective features for the speaker verification task or modeling the  other (O) information from the input.

The updated audio utterance is represented as $\boldsymbol{X^1_n} = x_1, x_2, \ldots, x_t$ and $\boldsymbol{X^2_n} = x_1, x_2, \ldots, x_t$, along with their corresponding pseudo labels (codewords) denoted as $\boldsymbol{Y^1_n} =  y_1, y_2, \ldots, y_t$ and $\boldsymbol{Y^2_n} = y_1, y_2, \ldots, y_t$. The total number of distinct pseudo labels (codewords) is denoted by $C$. For simplification we assume that the updated batch is twice the initial size ($2n$) and treat $\boldsymbol{X^1_n} == \boldsymbol{X^2_n} == \boldsymbol{X_n}$. 

During a forward pass, an audio utterance, $\boldsymbol{X_n}$, is fed to the CNN encoder. Approximately 50\% of the CNN encoder output tokens are masked using the $[M]$ embedding. To model the other information, $[USP]$ token embedding is appended at the beginning of the transformer encoder input. The updated input is passed to the transformer encoder, consisting of $l$ layers, resulting in hidden representations as $\boldsymbol{H_n} = h_{USP}^{1, 2, \ldots, l}, h_1, h_2, \ldots, h_t$. Where $h_{USP}^{1, 2, \ldots, l}$ and $h_1, h_2, \ldots, h_t$ embeddings are used to calculate the loss to model the other (O) and content (C) information respectively.

\vspace{1em}
\noindent \textbf{Content: Masked Prediction Loss}: 
Similar to HuBERT \cite{hsu2021hubert}, the MPL is computed by maximizing the cosine similarity of the hidden representation with the correct pseudo labels (codewords) and by minimizing with the incorrect ones as shown in Equation \ref{eq:MPL}, only at the masked indices.

\begin{equation}
\label{eq:MPL}
L_{MPL} = \frac{1}{\mathcal{M}} \sum_{i=1}^{M} \frac{ exp \ (\text{sim}(Ah_i, \mathbf{e}_c) / \tau )} {\sum_{c'=0}^{C-1} exp \ (\text{sim}(Ah_i, \mathbf{e}_{c'}) / \tau )}
\end{equation}

where $\mathcal{M}$ are the masked indices, $A$ is the projection matrix, $e_c$ and $e_{c'}$ are the correct and incorrect embeddings for the pseudo labels, $sim(·, ·)$ computes the cosine similarity between two vectors, and $\tau$ scales the logits.

\vspace{1em}
\noindent \textbf{Other: Utterance Similarity Prediction (USP) Loss}: 
Given USP token embeddings from $l$ transformer layers, $h_{USP}^{1, 2, \ldots, l}$. (i) To obtain a single embedding of other information for an utterance, all the USP embeddings from all the transformer layers are passed through the weighted layer function (WLF) i.e., $W_n = WLF( h_{USP}^{1, 2, \ldots, l} )$.(ii) Next, for either a key or query, the output from the WLF function is passed through two MLP layers similar to the \cite{chen2020simplesimclr}
followed by 2 MLP layers to avoid mod collapse.
Any two audio utterances within the batch are labeled as "Issame" if they originate from the same source and any other combination of two utterances is labeled as "Isnotsame", denoted by $\boldsymbol{S} \subseteq ({0, 1})$ or $(\textit{Isnotsame},\textit{Issame})$. 

The proposed $L_{USP}$ loss calculates the AMSoftmax loss \cite{wang2018additiveamsoftmax} as shown in Equation \ref{eq:USPloss}. The linear function maps any input of size 1. 

\begin{multline}
\label{eq:USPloss}
L_{USP} = \sum_{i=1}^n \sum_{j=1}^2 AMSoftmax\ ( \\
linear\ (\ concatenate\ [\ \boldsymbol{\Lambda}(X^j_i),\ \boldsymbol{\Lambda}(X^j_i)\ ]\ )\ ,\ S_{ij}\ \\
) 
\end{multline}
\textbf{W}here: 
\begin{align}
\label{eq:where1}
\hspace{0.1cm}  \boldsymbol{\Lambda} = MLP_2\ (\ MLP_1\ (\ WLF( h_{USP}^{1, 2, \ldots, l} ))) 
\end{align}

\vspace{1em}

In a batch, only fixed number of "Issame" pairs can be formed, exactly half, because the utterances are divided in two to form the pair. But the "Isnotsame" pairs can be formed within the batch or from the previous batches. Since the batch size in the \cite{audiodivide} paper is big compared to HuBERT, we use the negatives from the previous batches.

\vspace{1em}
\noindent \textbf{Other: Regularization Loss}: Two regularization (Reg)losses are proposed. (i) Similar to the SimCLR paper \cite{chen2020simplesimclr}, we use the normalized temperature-scaled cross entropy loss to maximize the similarity between the output of the $WLF$ embeddings of the positive pairs and vice-versa. (ii) We minimize the inverse of the similarity between the $h_{USP}^{l}$ and the average of $Avg( h_1^{l}, h_2^{l}, \ldots, h_t^{l} )$ to ensure that the information learned in the USP token embedding is distinct from that in the rest of the tokens in the transformer encoder layers. The resulting regularization loss is shown in Equation \ref{eq:Regloss}.

\begin{equation}
\label{eq:Regloss}
\small{
L_{Reg} = L_{SimCLR} + \sum_{i=1}^l \frac{1}{sim\ (\ h_{USP}^{i}\ , Avg( h_1^{i}, h_2^{i}, \ldots, h_t^{i}\ )}
}
\end{equation}

\vspace{1em}
\noindent \textbf{Total Loss}:
The total loss, denoted as $L$ is defined in Equation \ref{eq:combinedloss}. The goal is to maximize the representation's ability to jointly model the other (O) and content (C) information.

\begin{equation}
\label{eq:combinedloss}
L =  L_{MPL} + \alpha * (L_{USP} + L_{Reg})
\end{equation}

\section{Experimental details}
\label{section:experimentdetails}
The proposed O-HuBERT model closely mirrors the configuration of HuBERT base model \cite{hsu2021hubert} except few minor changes. It comprises a CNN encoder followed by 12 transformer encoder layers. Each transformer encoder layer consists of 768-dimensional hidden states and is equipped with 8 attention heads. The readers should keep in mind, that due to computational constraints in an academic setting, we only use the smallest possible model and dataset in this study to ensure a fair comparison with the base models in the literature. 

\vspace{1em}
\noindent \textbf{Pre-training}: 
Unlike HuBERT base, O-HuBERT has one $WLF$ and two $MLP$ layers to model the other information. This results in a total parameter count of 96.18 million, representing an increment of around 1.33 million parameters compared to HuBERT and these additional parameters are discarded after the pre-training step \footnote{The pre-training time does not differ much compared to HuBERT.}. O-HuBERT is trained for 400,000 iterations on 32 GPUs with a batch size of at most 87.5 seconds of audio per GPU. The best model checkpoint is determined using the dev-other subset. \footnote{Pre-trained models and training configurations will be made available after the acceptance.} 

\vspace{1em}
\noindent \textbf{Pre-training dataset}: The ASR Librispeech benchmark dataset \cite{panayotov2015librispeech}, which is derived from the LibriVox project, is used for pre-training. It has 3 splits (i) Training, comprising train-clean-100, train-clean-360, and
train-other-500, (ii) Development including dev-other and dev-clean, and (iii) Testing consists test-other and test-clean. Each data instance comprises an audio and its corresponding transcript.

\vspace{1em}
\noindent \textbf{Robust two-stage Data augmentation}: 
We randomly select half of the utterances in a batch and apply utterance mixing, similar to the UniSpeech-SAT \cite{chen2022unispeechsat, chen2022wavlm}. To the remaining utterances, we apply utterance mixing followed by reverberation similar to \cite{audiodivide}. Two stage data augmentation is shown to be very effective for solving speaker (other) based tasks \cite{xia2021selfaudiofivide2,zhang2021contrastiveaudiodivide4,audiodivide,huh2020augmentationaudiodivide3} and to avoid overfitting problem during modeling the other information.

\vspace{1em}
\noindent \textbf{Loss function}: 
In our proposed total loss as described in Equation \ref{eq:combinedloss}, $\alpha$ is set to 10. The linear function defined in Equation \ref{eq:where1} maps the concatenated output of 1536-dimension to a 1-dimensional space for binary classification. 
Negatives from previous batches are used to avoid overfitting in USP, SimCLR, and Reg loss. To ensure diversity among the negatives, we maintain a dictionary of negatives from a current batch and randomly select 1024 of negatives. To keep the training stable, negatives are used after the first 50k iterations.

\vspace{1em}
\noindent \textbf{SUPERB Benchmark}: 
The main objective of this study is (i) to investigate the effectiveness of O-HuBERT model to encode other (O) and content information using separate learnable parameters and (ii) effectiveness of robust data augmentation on achieving SOTA performance when compared to various downstream tasks. 
In order to comprehensively analyze the other and content information learnt by the O-HuBERT model we use several tasks from the SUPERB \cite{yang2021superb} benchmark and compare our method with other SOTA methods of similar model size. 

The SUPERB benchmark is designed to evaluate models on their capability to learn comprehensive audio characteristics on various downstream tasks spanning five aspects of the speech: speaker , content , semantics , para-linguistics , and generation . These downstream tasks are: (i) Three speaker tasks:- Speaker identification (SID), Automatic speaker verification (ASV), and speaker diarization (SD) (ii) Four content tasks:- Phoneme recognition (PR), Automatic speech recognition (ASR), Keyword spotting (KS), and Query-by-Example (QbE) (iii) Two Semantic tasks:- Intent classification (IC) and slot filling (SF) (iv) One para-linguistic task: Emotion recognition (ER) (v) Three generative tasks:- Speech enhancement (SE), Speech separation (SS), and Voice conversion (VC). For more details, refer \cite{yang2021superb}.

\vspace{1em}
\noindent \textbf{Other vs Content information}:
To better understand the learned other and content embeddings, we experiment with three different configurations: 

\begin{itemize}
    \item Using only the other information represented by the USP token embeddings $h_{USP}$. Given that the other information is Global in nature, we refer to this setup as global (G).
    
    \item  Using only the content information token embeddings $h_1, h_2, \ldots, h_t$. Given that the content information is local in nature, we refer to this setup as local (L). 

    \item Combined other and content information embeddings. To achieve this, weighted other embedding is added to all the content embeddings i.e., $\alpha*h_{USP} + h_1, \alpha*h_{USP} + h_2, \ldots, \alpha*h_{USP} + h_t$. The weights are learned during fine-tuning. We call this setup as GL. 
    
\end{itemize}


\section{Results}
\label{section:results}

In this section, we provide qualitative and quantitative result for our claims: (i) on the importance of modeling of other information using separate learnable parameters using three configurations. These configurations are denoted as G (other information only), L (content information only), and GL (combined other and content information). The GL configuration almost always performs better than either G or L on the SUPERB benchmark. (ii) On the importance of Robust two-stage data augmentation in achieving SOTA on the SUPERB benchmark.

\subsection{Importance of Modeling Other Information using separate learnable parameters}
\label{sec:importance of modelling g and l separately}

\begin{figure}[ht]
\centering
\scalebox{0.8}{
    \begin{tabular}{cc}
    
      \includegraphics[trim=0.55cm 4.1cm 2cm 4.5cm, clip,width=1.0\columnwidth]{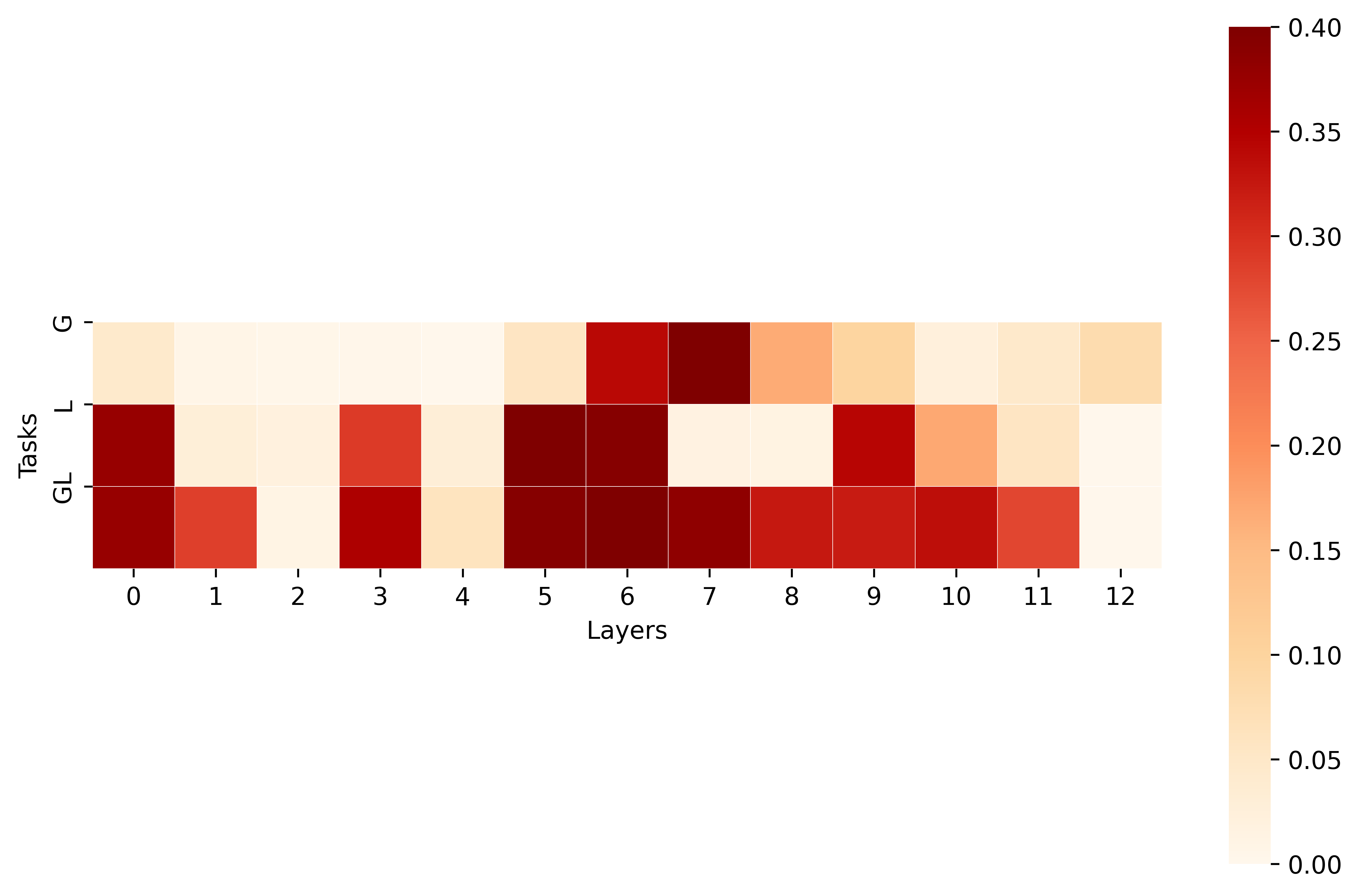} & \multirow{5}{*}{\includegraphics[trim=18cm 0cm 0cm 0cm, clip,width=0.09\columnwidth]{images/g-l-gl/SID_g-l-gl.png}}
      \\
      SID &  \\

      \includegraphics[trim=0.55cm 4.1cm 2cm 4.5cm, clip,width=1.0\columnwidth]{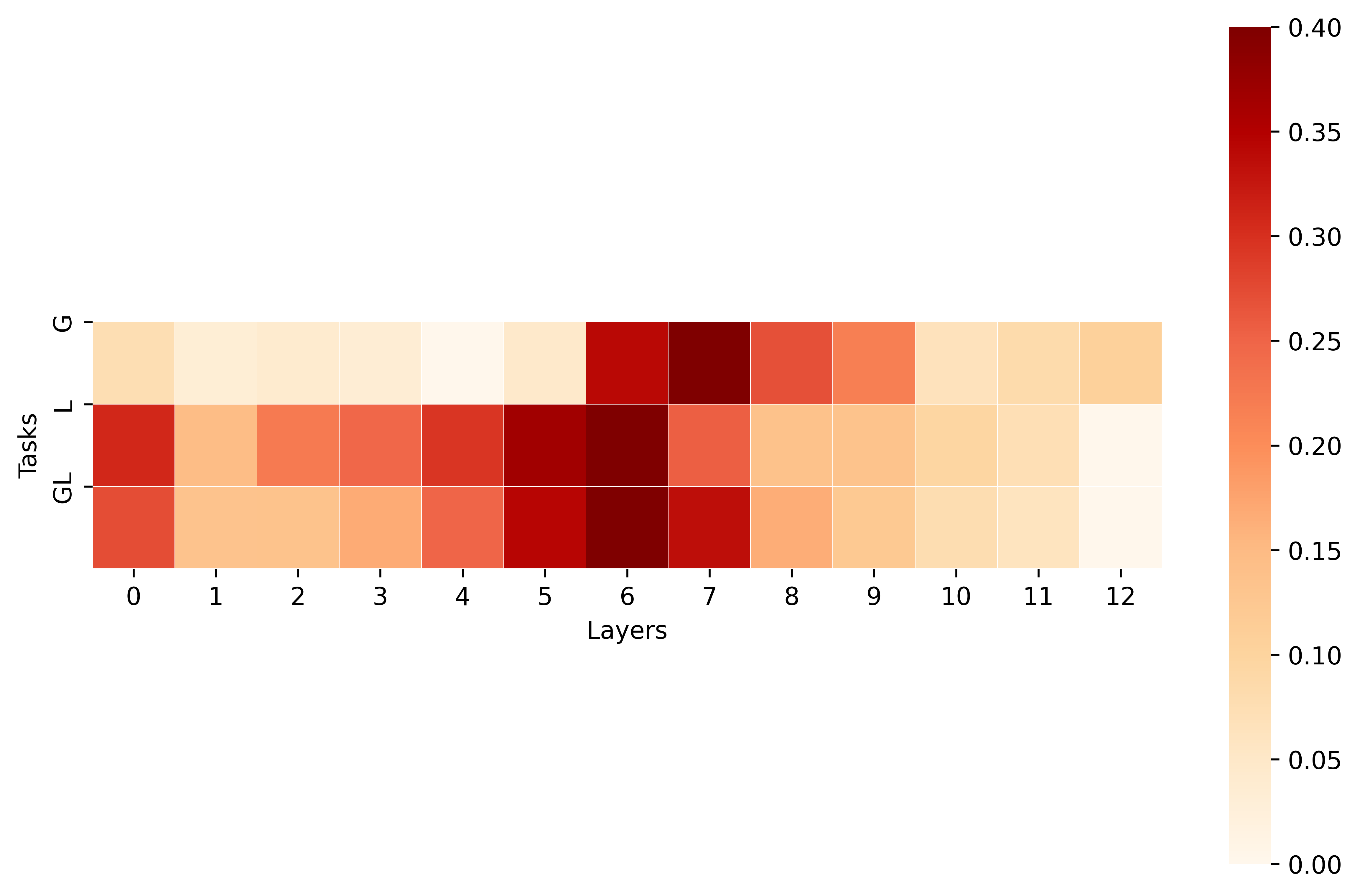} & \\
      ASV &  \\

      \includegraphics[trim=0.55cm 4.1cm 2cm 4.5cm, clip,width=1.0\columnwidth]{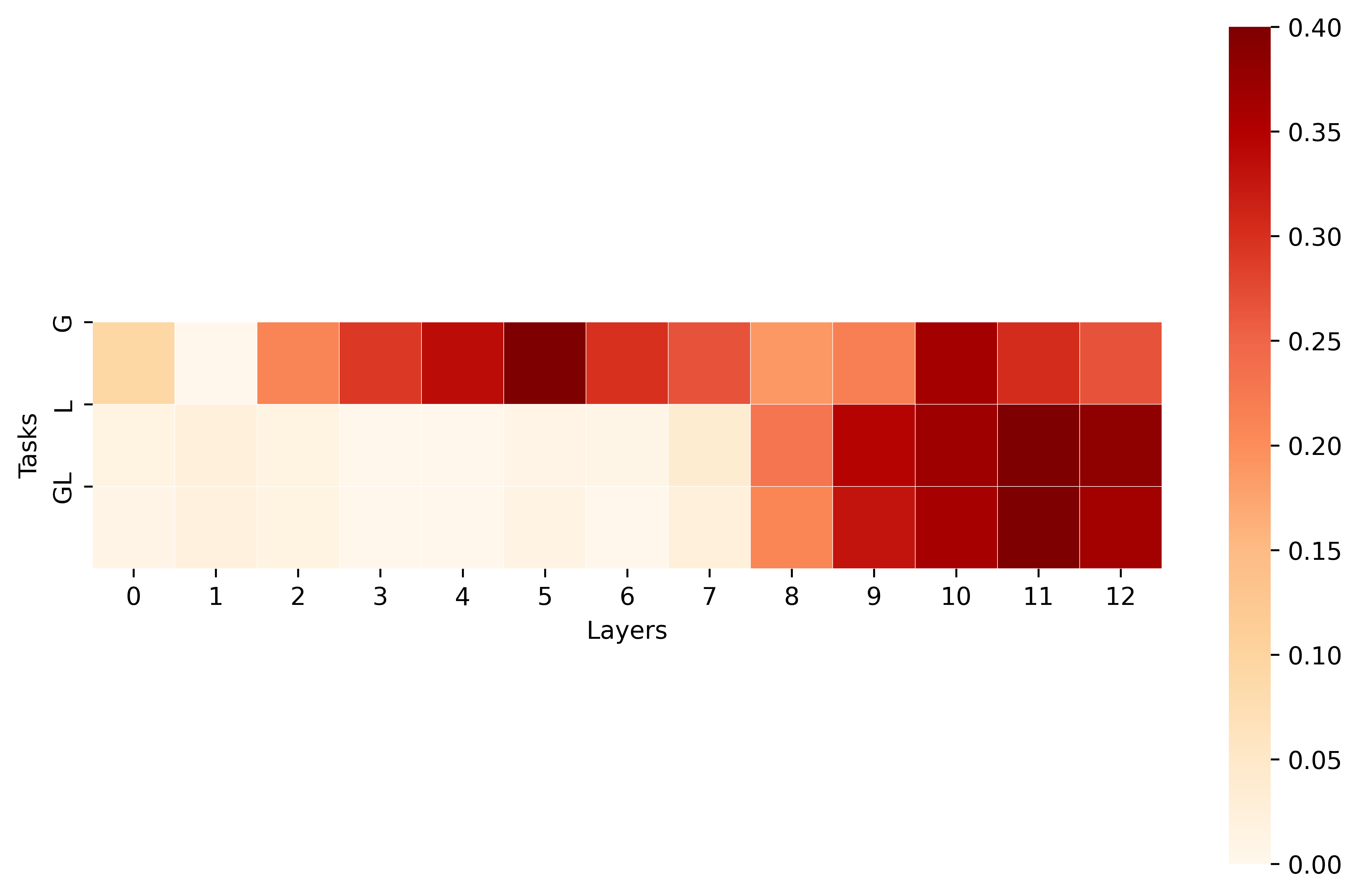} & \\
      IC & \\

        \end{tabular} 
        }
\caption{Importance of each transformer encoder layer (x-axis) vs different tasks for three configurations denoted as (i) G (other information only), (ii) L (content information only), and (iii) GL (combined other and content information).}
\label{figure:qualtitative-weight-analysis}
\end{figure}

\noindent \textbf{Qualitative analysis}:
Figure \ref{figure:qualtitative-weight-analysis} shows the importance of each transformer encoder layer for solving three downstream task in the SUPERB benchmark i.e., SID, ASV, and IC for three different configurations (G, L, and GL). 

In the O-HuBERT model, for the speaker based tasks, it is evident that later layers are more important in the G configuration compared to the L setup. This shows that the model is able to build high level complex features by effectively utilizing all the layers as shown in Figure \ref{figure:qualtitative-weight-analysis}. On the contrary, in the HuBERT and WavLM models only initial layers were shown to be important for solving tasks requiring speaker information (Other) i.e., SID and ASV. 
Lastly the GL setup combines both low level and high level features as evident from the increased participation of layers and the respective performance gains on the speaker based tasks as shown in Table \ref{table:GLGLR}. This suggests the importance of modeling other information using separate learnable parameters as done by the O-HuBERT model.

\renewcommand{\arraystretch}{1.2}
\begin{table}[ht]
\centering
\caption{ Comparison of the G, L, and GL setup on various speech downstream task from the SUPERB benchmark. * For the SID task, the test accuracy is higher for the L setup compared to GL setup is pure coincidence/luck given that the dev accuracy is not reflecting the same trend as shown in Figure \ref{figure:weightanalysis}. Furthermore the GL setup is more stable compared to the L setup as shown by the dev and test accuracy plots. Based on these two observations, we hypothesize that a thorough hyper-parameter search on the learning rate could give better results for the GL setup compared to the L setup.
} 
\label{table:GLGLR}
\scalebox{1}{
\begin{tabular}{|p{1.5cm}|p{1cm}|p{1cm}|p{1cm}|p{1cm}|}
\hline
    Task & G & L & GL & Random  \\
    \hline
    \hline
    SID & 75.6 & 86.21$^*$ & 85.35 & 14.7  \\
    \hline
    ASV & 6.54 & 4.56 & \textbf{4.26} & - \\
 
    \hline
    \hline
    PR & - & \textbf{5.52} & 5.55 & - \\
    \hline
    ASR & - &  7.09 & \textbf{6.97} & - \\
    
    \hline
    \hline
    ER & 60.87 & \textbf{68.55} & 68.24 & - \\

    \hline
    \hline
    SE & - & 2.65 & \textbf{2.67} & - \\
    \hline
    VC & - & 7.38 & \textbf{7.41} & - \\ 
    \hline
    
    \end{tabular}}
\end{table}

On the other hand, for the IC task this is not the case and there is no substantial change in the layer participation when using the L and GL setup. We hypothesize that (1) solving the IC task requires effective modeling of the content (local) information and therefore adding the other (global) information might not add any meaningful new information. Or (2) the performance on the IC task is already very high, close to 98\%.

\noindent \textbf{Quantitative analysis}:
As shown in Table \ref{table:GLGLR}, while the G setup, representing other information, does capture relevant details but the performance still lags behind the L setup. The bottleneck could be using a single USP token to encode the other information present in the whole utterance which may not be sufficient. Therefore a need arises to use multiple USP tokens, maybe one USP token per second of an audio to encode the other information effectively. We leave this for future work. 
Furthermore, the GL setup, representing combined other and content information, gains performance across most downstream tasks, except for the emotion recognition task. This shows that for the ER task on SUPERB, encoding content information is essential for high performance.

An equivalent of the G setup is features extracted at one random token from the L setup. When compared, the performance on the SID task degrades sharply when using the random setup. This shows that the other information learned in the USP token is not random and the O-HuBERT model is working as expected. We also fine-tuned the O-HuBERT encoder and found that G setup performs better compared to the L setup. This is because, averaging all the tokens in the L setup introduces instability during finetuning. 

The above results corroborate our claim on the importance of modeling other information using separate learnable parameters to utilize all the layers to build deep and complex high level features.

\renewcommand{\arraystretch}{1.3}
\begin{table*}[ht]
\centering
\caption{UNIVERSAL SPEECH REPRESENTATION EVALUATION ON SUPERB BENCHMARK.}
    \scalebox{0.57}{
    \begin{tabular}{|p{2cm}|p{0.9cm}|p{1.4cm}|p{0.9cm}|p{1cm}|p{1cm}|p{0.9cm}|p{1.05cm}|p{0.9cm}|p{1.37cm}|p{1cm}|p{1cm}|p{1cm}|p{1cm}|p{1cm}|p{1.105cm}|p{1.05cm}|p{1.5cm}|p{1.05cm}|p{1.05cm}|}
    \hline
      \multicolumn{1}{|c|}{\multirow{3}{*}{\textbf{Method}}} & \multicolumn{1}{c}{\multirow{3}{*}{\textbf{\#Params}}}  &  \multicolumn{1}{|c|}{\multirow{3}{*}{\textbf{Corpus}}} &\multicolumn{3}{|c|}{\textbf{ Speaker}} & \multicolumn{4}{c|}{\textbf{Content}} & \multicolumn{3}{c|}{\textbf{Semantics}} & \textbf{ParaL} & Score & \multicolumn{5}{|c|}{\textbf{Generation}} \\
      \cline{3-20}
       & & & SID & ASV & SD & PR & ASR & KS & QbE & IC & \multicolumn{2}{|c|}{SF} & ER & & \multicolumn{2}{|c|}{SE} & SS & \multicolumn{2}{|c|}{VC} \\
       \cline{3-20}
        & & & Acc $\uparrow$ & EER $\downarrow$ & DER $\downarrow$ & PER $\downarrow$ & WER $\downarrow$ & Acc $\uparrow$ & MTWV $\uparrow$ & Acc $\uparrow$ & F1 $\uparrow$ & CER $\downarrow$ & Acc  $\uparrow$ &  & PESQ $\uparrow$ & STOI $\uparrow$ & SI-SDRi $\uparrow$ & MCD $\downarrow$ & WER $\downarrow$\\
        \hline
        \hline

        FBANK & 0 & - & 8.5e-4 & 9.56 & 10.05 & 82.01 & 23.18 & 8.63 & 0.0058 &  9.10 & 69.64 & 52.94 & 35.39 & 0 & 2.55 & 93.6 & 9.23 & 8.47 & 38.3\\
        \hline

        modified CPC & 1.84M & LL 60k hr & 39.63 & 12.86 & 10.38 & 42.54 & 20.18 & 91.88 & 0.0326 & 64.09 & 71.19 & 49.91 & 60.96 & 278  & 2.57 & 93.7 & 10.40 & 8.41 & 26.2 \\ 
        \hline
        HuBERT Base & 94.68M & LS 960 hr & 81.42 & 5.11 & 5.88 & 5.41 & 6.42 & 96.30 & 0.0736 & 98.34 & 88.53 & 25.20 & 64.92 & 941  & 2.58 & 93.9 & 9.36 & 7.47 & 8.0 \\ 
        \hline
        
        WavLM Base  & 94.70M & LS 960 hr & 84.51 & 4.69 & 4.55 & \textbf{4.84} & \textbf{6.21} & 96.79 & 0.0870 & \textbf{98.63} & \textbf{89.38} & \textbf{22.86} & 65.94 & 1019  & 2.58 & 94.0 & 10.37 & 7.42 & \textbf{8.0}\\ 
        
        \hline
    
        OHuBERT & 96.18M & LS 960 hr & \textbf{85.36} & \textbf{4.26} & \textbf{4.12} & 5.54 & 7.09 & \textbf{97.34} & \textbf{0.1082} & 98.21 & 88.64 & 24.38 & \textbf{68.55} & \textbf{1080}  & \textbf{2.67} & \textbf{94.18} & \textbf{10.42} & \textbf{7.38} & 12.52 \\
        
        
        \hline
        
        \end{tabular}}
\label{table:superb}
\end{table*}


\subsection{Importance of Robust two-stage Data Augmentation}

\noindent \textbf{Universal Representation Evaluation on the SUPERB benchmark}: 
When evaluated across a wide range of tasks on the SUPERB benchmark, our O-HuBERT model achieves a new SOTA with a comparable model size and pre-training data, as shown in Table \ref{table:superb}. 
Tasks such as Emotion Recognition (ER) and Query-by-Example (QbE) benefit the most from our two-stage data augmentation strategy, outperforming WavLM, as demonstrated in Table \ref{table:superb}. Additionally, speaker-based tasks also show improvements over WavLM, which uses only utterance mixing for data augmentation. These observations highlight the importance of robust data augmentation during pre-training, as opposed to the simpler utterance mixing strategy used in WavLM \cite{chen2022wavlm} and UniSpeech-SAT \cite{chen2022unispeechsat}, in achieving better features for tasks that require other information orthogonal to content information.

Generation tasks also benefit from the two-stage data augmentation.
Overall, the results indicate that the majority of speech downstream tasks require information orthogonal to the ASR task. This suggests that the research community should develop pre-training methods with an inherent bias toward Joint Optimization of Other and Content Information (JOOCI), rather than primarily focusing on content as is currently the norm.

\begin{figure}[ht]
\centering
\scalebox{0.48}{
    \begin{tabular}{cc}
      \includegraphics[width=1.0\columnwidth]{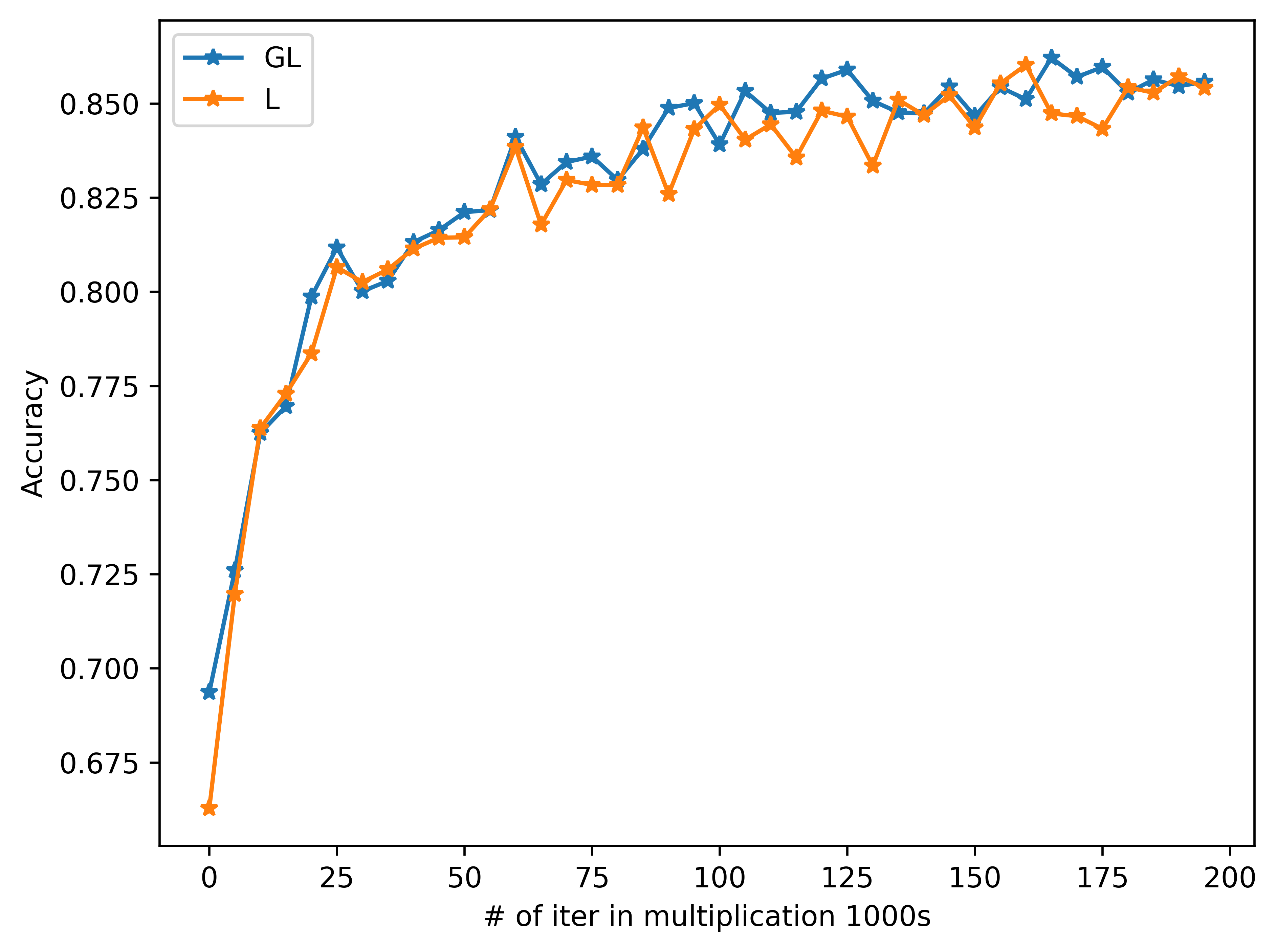} & 
      \includegraphics[width=1.0\columnwidth]{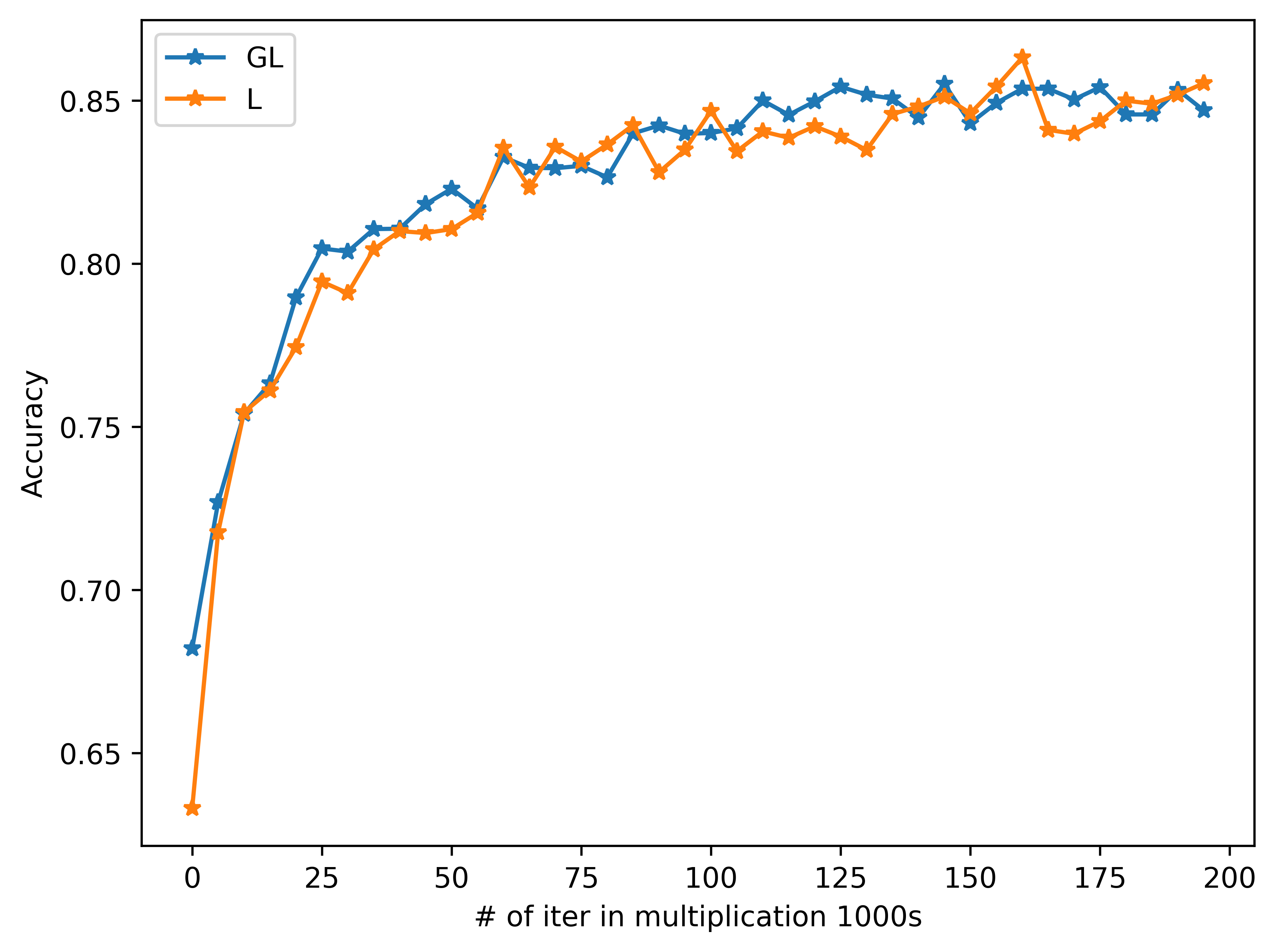} \\
        (a) dev & (b) test \\
        \end{tabular} 
        }
\caption{Comparison of the GL and L setup dev and test accuracy plots on the SID task from the SUPERB benchmark.}
\label{figure:weightanalysis}
\end{figure}

\subsection{Ensemble}
Using the output layers embedding from one forward pass of the O-HuBERT model, we train three different classifiers using the G, L, and GL configuration for the SID task. 
Table \ref{table:ensemble_top5} show the ensemble results. 
We observe a direct relationship in between the performance gains for different tasks and  the number of classes. This makes sense, because ensemble helps in reducing the ambiguity with voting. For the SID task, the gain is of approximately 5\%. For related works, in the SUPERB benchmark a similar parameter model (WavLM+) pre-trained using 90,000 hours of audio data achieves an accuracy of 89.42\%. In contrast, our method, pre-trained on 960 hours and coupled with an ensemble of three classifiers, achieves a higher accuracy of 89.78\%

\renewcommand{\arraystretch}{1.1}
\begin{table}[ht]
\caption{Comparison of the different tasks from the SUPERB benchmark when using ensemble. IC task is special because it has three set of sub-classes for each input i.e., 6,10,14 classes for each class.}
\centering
\scalebox{0.9}{
    \begin{tabular}{|c|c|c|c|c|c|c|}
    \hline
        Task & SID & KS & PR & ASR & IC & ER\\
        \hline
        \hline
        - & 85.36 & 97.34 & 5.54 & 7.095 & 98.21 & 68.55 \\
        Ensemble & 89.78 & 97.47 & 5.52 & 7.093 & 98.21 & 68.55  \\
        \hline
        \# of classes & 1251 & 12 & 74 & 32 & - & 4 \\ 
        \hline
        \end{tabular}}
\label{table:ensemble_top5}
\end{table}

\section{Related Work}
\label{section:related works}

Different speech tasks require effective modeling of either content information, other information, or both. Content information tends to be more detailed and specific compared to other information. For instance, a high-quality text-to-speech (TTS) system must accurately model both content and other information (such as style), as shown in previous studies like \cite{huang2022generspeech, ren2019fastspeech, ren2022fastspeech2, shen2023naturalspeech2, ju2024naturalspeech3}. The literature also indicates that encoding speech at different frequencies is advantageous for various tasks, including speech recognition \cite{asr1,asr2,asr3,asr4,asr5,asr6}, speaker verification \cite{sv1}, speech enhancement \cite{SE1}, and voice conversion \cite{VC1}.

Models that utilize masked predictive loss (MPL) during pre-training, either partially or entirely \cite{baevski2020wav2vec2.0, chung2021w2vbert, hsu2021hubert}, tend to excel in content-based tasks like ASR. Moreover, research by \cite{shi2023explorationMRHUBERT, shi2023MRHuBERT, yadav2023analysing} has shown that applying MPL at multiple resolutions enhances performance on content-based tasks such as ASR, though it comes at the expense of reduced performance in speaker-related tasks. Ideally, an ASR system should focus solely on encoding what is being said (content) and disregard everything else (other). This challenge arises because content and other information are orthogonal, making joint optimization difficult, as demonstrated by \cite{yadav2023analysing, yang2021superb}. Consequently, there is a need to use separate learnable parameters to model content and other information effectively.

Data augmentation is essential to model other information as shown by \cite{chen2022unispeechsat} and robust two-stage data augmentation has been shown to be highly effective for speaker-based tasks \cite{xia2021selfaudiofivide2, zhang2021contrastiveaudiodivide4, audiodivide, huh2020augmentationaudiodivide3}. The loss function is also equally important as  \cite{audiodivide, wang2018additiveamsoftmax, mohamed2022self}.

\section{Conclusion and future work}
\label{section:conclusion}
In this work, we introduced Other HuBERT (O-HuBERT) model, a modified version of the HuBERT model, specifically designed to model other information using separate parameters, allowing for more complex features by utilizing all layers. In contrast to traditional self-supervised learning models. Furthermore, we also show that the majority of speech downstream tasks require information orthogonal to
the ASR task and benefit from robust data augmentation. 
This suggests that the research community should develop pre-training methods which are inherently capable of Joint Optimization of Other and Content information (JOOCI), rather than focusing primarily on content, which is the current norm.

\noindent The lessons from this study highlight the potential of the O-HuBERT method; however, it still faces two key challenges that need to be addressed and we leave this for future work:

\begin{enumerate}
    \item The current single USP token approach struggles to adequately capture and encode all other information, leading to sub-optimal performance on tasks like SID, ASV, and ER. Addressing this requires devising methods to incorporate other information at higher resolutions i.e., using multiple $USP$ tokens and not just one.

    \item There's a need to design pre-training methodology which is capable of modeling content information without being affected by data augmentation. Though Robust data augmentation is important for a various speech tasks it harms the modelling of content information as shown by our results and previous studies. That is to say, methods which can jointly optimize the other and content information, without one affecting the other. This will help in improving the performance on the content based tasks such as ASR and PR. 
    
\end{enumerate}

\bibliography{aaai25}

\end{document}